\documentclass{article}

\usepackage{arxiv}

\usepackage[utf8]{inputenc} 
\usepackage[T1]{fontenc}    
\usepackage{hyperref}       
\usepackage{url}            
\usepackage{booktabs}       
\usepackage{amsfonts}       
\usepackage{nicefrac}       
\usepackage{microtype}      
\usepackage{cleveref}       
\usepackage{lipsum}         
\usepackage{graphicx}
\usepackage{natbib}
\usepackage{doi}
\usepackage{subcaption}
\usepackage{graphicx}
\usepackage{multirow}
\usepackage{lipsum} 
\usepackage{tcolorbox} 

\AtBeginDocument{%
  \providecommand\BibTeX{{%
    \normalfont B\kern-0.5em{\scshape i\kern-0.25em b}\kern-0.8em\TeX}}}

\title{Clustered Retrieved Augmented Generation (CRAG)}

\date{}

\author{ \hspace{1mm}Simon \.{A}kesson\\
	AI Center of Excellence\\
	Sinch AB\\
	Stockholm, Sweden \\
	\texttt{simon.akesson@sinch.com} \\
	\And
	\hspace{1mm}Frances A. Santos \\
	AI Center of Excellence\\
	Sinch AB\\
	Campinas, Brazil \\
	\texttt{frances.santos@sinch.com} \\
}

\renewcommand{\shorttitle}{\.{A}kesson and Santos}

\hypersetup{
pdftitle={Clustered Retrieved Augmented Generation (CRAG)},
pdfsubject={cs.AI},
pdfauthor={Simon \.{A}kesson, Frances A. Santos},
pdfkeywords={Clustering, Summarization, Generative AI, Retrieved Augmented Generation, Question-Answering Systems},
}

\begin{document}
\maketitle

\begin{abstract}
  Providing external knowledge to Large Language Models (LLMs) is a key point for using these models in real-world applications for several reasons, such as incorporating up-to-date content in a real-time manner, providing access to domain-specific knowledge, and contributing to hallucination prevention. The vector database-based Retrieval Augmented Generation (RAG) approach has been widely adopted to this end. Thus, any part of external knowledge can be retrieved and provided to some LLM as the input context. Despite RAG approach's success, it still might be unfeasible for some applications, because the context retrieved can demand a longer context window than the size supported by LLM. Even when the context retrieved fits into the context window size, the number of tokens might be expressive and, consequently, impact costs and processing time, becoming impractical for most applications. To address these, we propose CRAG, a novel approach able to effectively reduce the number of prompting tokens without degrading the quality of the response generated compared to a solution using RAG. Through our experiments, we show that CRAG can reduce the number of tokens by at least 46\%, achieving more than 90\% in some cases, compared to RAG. Moreover, the number of tokens with CRAG does not increase considerably when the number of reviews analyzed is higher, unlike RAG, where the number of tokens is almost 9x higher when there are 75 reviews compared to 4 reviews.
\end{abstract}


\section{Introduction}
The vector database-based Retrieval Augmented Generation (RAG) approach, proposed in \cite{lewis2020retrieval}, has been widely adopted for Question-Answering (QA) systems using LLMs \cite{anand2023sciphyrag, wang2023augmenting, mansurova2023development}. In such systems, we can index all relevant content regarding some topic into a knowledge base (i.e., the vector database). Then, given a question, the RAG method retrieves from the knowledge base the relevant knowledge that is provided to LLM for answer generation. In this way, by combining RAG and LLMs, we can bypass the context window size limitation and, consequently, reduce the costs of using LLMs, since the number of input and generated tokens have a direct impact on computational resource usage. Moreover, RAG helps to address other critical limitations for LLMs, such as incorporating up-to-date content from external sources and providing access to domain-specific knowledge. 

Besides QA systems, we can also apply RAG and LLMs together to enable many other applications, such as semantic search, personal assistants, and creative content creation. In common, such applications segment and provide LLM with a part of knowledge indexed into the knowledge base, which is sufficient for the LLM to generate the expected response. However, in some applications, the LLM would need to consult the whole knowledge base to generate the response, which surely would extrapolate the context window size. Consider the following example to illustrate this. 

A big retail company wants to understand its customers' opinions better about the products offered by them. To this end, they decided to collect all reviews shared by their customers in the past year for every product, resulting in hundreds of thousands of reviews written in natural language. Then, they use a vector database for storage of all reviews, in a way they can use some LLM to answer some questions related to the product reviews, helping them to extract relevant insights from reviews. However, they still need to figure out how to connect the vector database to the LLM, since all product reviews should be analyzed by LLM to generate the right answer, representing a huge amount of input tokens, which makes RAG a non-viable option.

\begin{figure}
  \centering
  \includegraphics[scale=0.5]{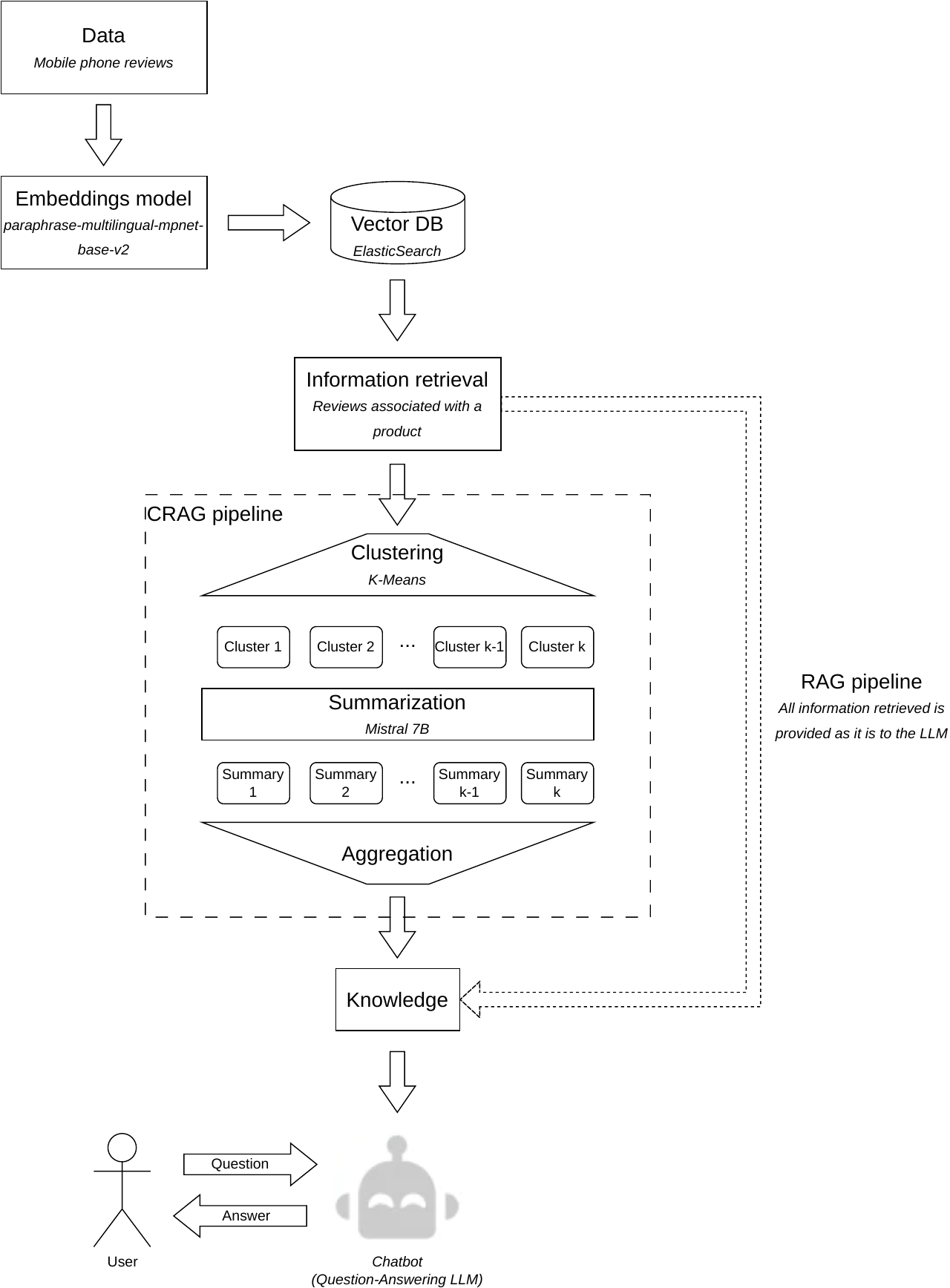}
  \caption{Overview of CRAG-pipeline helping a Question-Answering system with external knowledge.}
  \label{fig:crag}
\end{figure}

For that, we propose a Clustered Retrieved Augmented Generation (CRAG) approach, as shown in Figure \ref{fig:crag}. As we can see, CRAG alters the pipeline that normally constitutes RAG, including three subsequent steps: clustering, summarization, and aggregation. First, CRAG splits the reviews into $k$ clusters, where $k$ must be informed as an input parameter. Then, for every cluster obtained, CRAG summarizes the cluster content to generate a brief review that has the main information that represents the entire cluster. Finally, all summaries are merged into a single text to create the knowledge generated by CRAG. 

We show that CRAG can effectively reduce the number of tokens by at least 46\%, achieving more than 90\% in some cases, compared to the RAG approach, without degrading the quality of the response generated by Question-Answering Chatbot, regardless of LLM used to this end. By doing that, CRAG contributes to cost reductions, fits external knowledge in the model context window size, and reduces the model's latency to generate responses.

\section{Methodology}

\subsection{Data} \label{subsec:data}
Considering the example of a big retail company (described in the previous section), we found a publicly available dataset on the Kaggle\footnote{https://www.kaggle.com/datasets/nehasontakke/amazon-unlocked-mobilecsv. Last accessed on February $3^{rd}$, 2024.} platform related to mobile phones bought on Amazon. This dataset has 413,841 entries, where each entry consists of the following fields: product name, brand name, price, rating (ranging from 1 to 5), reviews, and review votes (a counter indicating how many users vote in the review). 

We only consider the product name and the reviews associated with them in our study. All information is written in English. In all, there are 4,410 distinct products and 414K reviews about them, around 94 reviews per product on average. The product with the largest number of reviews has 1,451 entries, which would require 35,787 tokens for the input to GPT-4. Some reviews are too generic, e.g., ``Very pleased'', favoring duplicated cases. Nevertheless, there are more than 162K unique reviews for all products.


\subsection{Embeddings}
Given the reviews, the next step is to generate numerical representations of the reviews, named embeddings, to store them in a vector database. To this end, we used a sentence transformer model called \emph{paraphrase-multilingual-mpnet-base-v2} \cite{reimers-2019-sentence-bert}. This model is a version of MPNet pre-trained on downstream tasks for paraphrasing across 50 languages. As MPNet architecture combines the benefits of BERT architecture, which uses Masked Language Modeling (MLM), and XLNet, which uses Permuted Language Modeling (PLM), it might cancel out what either model lacks \cite{song2020mpnet}. Such capability makes it a suitable choice to group sentences as it works well for the semantic textual similarity task \cite{reimers2019sentencebert}. By applying the Paragraph embeddings to the reviews, each review is mapped into 768-dimensional dense vector space, where similar reviews are also mapped close, enabling us to measure their similarities according to their distance in the vector space.

\subsection{Clustering}
Having the embeddings for all reviews, we can group similar reviews for the same product using some clustering algorithm. Several techniques can be used to perform clustering, where one of the most famous methods is K-means, given its simple and intuitive operation \cite{tan2018introduction}. 

K-means is an unsupervised machine learning algorithm, where we must choose the number of clusters, denoted by the parameter $k$, before starting. Then, initial centroids are defined for each of the $k$ clusters in a multidimensional space, which may be defined randomly or by some heuristic \cite{macqueen1967classification}. After that, each object is allocated to the cluster whose centroid is the most similar to it (i.e., nearest centroid) \cite{macqueen1967classification}. When done, the position of centroids is adjusted based on the mean value of distances for each cluster, and then, the process is repeated until the algorithm converges (i.e., data points stay in the same cluster), or reaches a stopping criterion (a predefined maximum number of iterations) \cite{macqueen1967classification}. 

To define a suitable value of $k$, we consider the elbow method, ranging from 1 to 10, and calculate the Inertia (sum of squared distances of samples to their closest cluster center) to estimate how well a dataset was clustered by K-Means. After empirical testing using the reviews, we defined $k=4$, despite bigger values of $k$, such as 5 and 6, which could be considered as well, however, we opted to keep the smaller candidate. 

\subsection{Summarization and Aggregation}
For every product in the dataset, we separated its reviews into four clusters, where each cluster has reviews with semantic similarity among them. For instance, by exploring some clusters, we found dozens of reviews mentioning the same thing about the same product, which was compelling evidence we could reduce the number of tokens.  

For that, we chose Mistral 7B \cite{jiang2023mistral}, a 7-billion-parameter open-source language model, for summarizing the clusters' content. Mistral 7B has an impressive performance in several benchmarks, outperforming other prominent and larger competitors, such as LLaMA-13B \cite{jiang2023mistral}. Mistral 7B is based on Transformer architecture and covers (NLP) tasks such as text, code, and mathematics generation. 

Thus, we use one-shot prompting, where one example was taken from the dataset and processed according to the desired output, to generate summaries with Mistral 7B. By using zero-shot prompting, we could not yield satisfactory results. In some cases, more than one example should be provided to the model (called few-shot prompting), depending on the complexity of the task. In our case, using only one example was sufficient. See below the one-shot prompting used to summarize the content of the clusters using Mistral 7B:

\begin{small}
    
\bigbreak
\noindent
\textbf{[INST] Given a series of reviews, create a concise summary that effectively conveys the overall sentiment and key themes without directly quoting the reviews. Focus on distilling the main ideas and emotions expressed in the reviews, providing a clear and accurate representation of the conversation's tone and content. Do not reference the reviews.}

\noindent
Reviews: \{\{PRODUCT\_REVIEWS\}\}

\noindent
Question: \{\{ONESHOT\_QUESTION\}\} [/INST]

\noindent
Answer: \{\{ONESHOT\_ANSWER\}\}

\bigbreak
\end{small}

\noindent, where [INST] and [/INST] are special instruction tokens used by Mistral to delimit the instruction. \{\{PRODUCT\_REVIEWS\}\} denotes the collection of reviews for a given product. The question and answer provided to Mistral 7B as an example are denoted by \{\{ONESHOT\_QUESTION\}\} and \{\{ONESHOT\_ANSWER\}\}, respectively.

As a result, we obtained four summaries, one for each cluster, that represent the original content, i.e., the reviews of a product. We repeat this process, clustering, and summarization of reviews, for all products of the dataset. Thus, we filtered out repetitive content and joined similar ones, reducing the number of tokens significantly. Finally, for every product, we merge its summaries into a single text to create the knowledge generated by CRAG.

\subsection{Chatbot: Question-Answering LLM}
Similar to RAG pipelines, where external knowledge after preparation is used as input for an LLM to help the model generate the expected response, we also feed an LLM with knowledge prepared by CRAG. We chose three state-of-the-art LLMs, GPT-4-0613 model (or, GPT-4, for simplicity), Llama2-70B, and Mixtral8x7B, to create three different versions of Question-Answering (QA) Chatbot.

We use the following prompt to LLMs generate answers for users' questions based on external knowledge provided by CRAG or RAG:
\begin{small}
\bigbreak
\noindent
\textbf{You will be provided with a set of descriptions of messages. You will also be provided with a question. Given these descriptions, answer the question in 300 words. If applicable, apply examples to justify your answer. Answer in bullet points.}

\noindent
Related descriptions: \{\{KNOWLEDGE\}\}

\noindent
Question: \{\{USER\_QUESTION\}\}

\bigbreak
\end{small}

\noindent, where \{\{KNOWLEDGE\}\} indicate the knowledge produced by either CRAG or RAG. The \{\{USER\_QUESTION\}\} represents the question to be answered by QA Chatbot. In this way, we can interact with the chatbot to clarify any doubts related to reviews shared by customers.

\section{Experiments}
The experiments were implemented in Python 3.9 using PyTorch library, with the support of two NVIDIA GeForce RTX 3090 GPUs for embedding creation and summarization.

\subsection{Experimental setup}

\subsubsection{Dataset}
As described in Section \ref{subsec:data}, we adopt the dataset with users' reviews about mobile phones bought on Amazon. For products with less than four reviews, we filtered out from the dataset, because in such cases the number of tokens would not extrapolate the context window size of LLMs and, consequently, not require any further processing, making it possible to include all reviews in the prompt as they are.

\subsubsection{Comparisons}
We evaluate answers generated using CRAG's approach against a traditional RAG pipeline. To this end, we developed a RAG pipeline using ElasticSearch as a database. With ElasticSearch, we can effectively retrieve information needed. In this way, using the RAG approach, all reviews associated with a product are retrieved and used as input in the prompt for QA Chatbot, resulting in a considerable number of tokens in certain cases.

\subsubsection{Evaluation Metrics}
To measure CRAG's performance compared to RAG, we count the number of tokens in the prompt using CRAG and RAG, for the Question-Answering LLMs (i.e., GPT-4, Llama2-70B, and Mixtral8x7B). Since such LLMs use different tokenizers, slight changes in the number of tokens might happen and, therefore, we let T-CRAG and T-RAG denote the maximum number of tokens by using CRAG and RAG, respectively. Thus, we can assess which approach produces fewer tokens and, consequently, represents lower costs. We denoted by CiT (Change in Tokens) the percentage difference between T-CRAG and T-RAG, where negative percentages mean T-CRAG demands fewer tokens, and positive otherwise. Moreover, we also compare the semantic similarity between output generated using RAG and CRAG approaches, considering the different LLMs. In this way, we can evaluate if both approaches generate similar responses, despite differences in the number of tokens. For this, we consider the cosine similarity (denoted by CosSim) between the outputs of each pipeline (CRAG and RAG).

\subsection{Results}
\begin{table*}[ht]
  \centering
  \caption{Results obtained for 7 products with diverse number of reviews.}
  \begin{tabular}{ccccccc}
    \toprule
    \small{\textbf{\# reviews}} & \small{\textbf{T-CRAG} }& \small{\textbf{T-RAG}} & \small{\textbf{CiT(\%)}} & \small{\textbf{CosSim (GPT-4)}} & \small{\textbf{CosSim (Llama2-70B)}} & \small{\textbf{CosSim (Mixtral8x7B)}} \\
    \midrule
    4 & 259 & 607 & -57.33\% & 0.87 & 0.82 & 0.85\\
    17 & 404 & 758 & -46.70\% & 0.84 & 0.85 & 0.79\\
    30 & 422 & 1099 & -61.60\% & 0.91 & 0.66 & 0.75\\
    37 & 490 & 2145 & -77.16\% & 0.88 & 0.80 & 0.86\\
    73 & 466 & 5095 & -90.85\% & 0.77 & 0.70 & 0.84\\
    42 & 473 & 2945 & -83.93\% & 0.90 & 0.86 & 0.81\\
    75 & 468 & 5165 & -90.93\% & 0.88 & 0.76 & 0.74\\
    \bottomrule
  \end{tabular}
  \label{tab:results}
\end{table*}

As we can see in Table \ref{tab:results}, by using CRAG we achieve a significant reduction in the number of tokens. In most of the cases, CRAG was able to reduce the number of tokens by at least 46\%, achieving more than 90\% in some of them (probably when there exists more duplicated reviews). In addition to that, we can see that the number of tokens with CRAG does not increase considerably when the number of reviews increases. On the other hand, using RAG, the number of tokens is almost 9x higher when there are 75 reviews in comparison to 4 reviews. Considering that the pricing for OpenAI's API to use GPT-4 is \$0.01/1k prompt tokens \cite{openai2023gpt4}, for this case where RAG results in 5,165 tokens and CRAG only 468, we have a significant cost reduction from \$0.051 to \$0.004 per query, more than ten times cheaper. This means that over time, large-scale applications would have a huge cost reduction only replacing RAG by CRAG in their pipelines.

We can also see the output generated using both approaches is quite similar, where the CosSim values vary around 0.70, independently of the LLM. Moreover, the difference between outputs does not mean that one answer is worse, or better than another, it only shows there is a divergence between them. Looking at outputs produced by GPT-4 with both CRAG and RAG, we can see that they cover the same topics, and both do well, as the following example: 

\begin{itemize}
    \item \textbf{Generated response with CRAG:}
    \begin{itemize}
        \item Several users have had issues with the quality and durability of the holsters, often experiencing frequent breakage.
        \item Despite the durability issues, some customers compare these holsters favorably to Otterbox, noting similar appearance and functionality but at a lower price.
        \item There are instances of users having to get replacements or buy additional holsters because of the aforementioned breakage.
        \item Many customers appreciate the affordability and durability.
    \end{itemize}
    \item \textbf{Generated response with RAG:}
    \begin{itemize}
        \item Some users note that the product is not of the highest quality, with several people suggesting the holsters break easily. For instance, one user noted that their original Otterbox holster lasted over a year, but they broke all three of these replacement holsters within a few weeks.
        \item A few users say the product does its job adequately, but just barely.
        \item Many users purchase the holsters for work, stating they are durable and cost-effective.
    \end{itemize}
\end{itemize}

From this example, we can see that responses generated roughly state the same thing, i.e., some users had issues with the durability, while others were satisfied with it. Affordability was brought up in both as well.  

Therefore, our results indicate that CRAG can effectively reduce the number of tokens, which contributes to cost reductions, fits external knowledge in the model context window size, and reduces the response latency (since the model needs to process fewer input tokens), at the same time it does not degrade the quality of response generated compared to a solution using RAG.

\section{Conclusion and Future work}
We propose CRAG, a novel approach able to effectively reduce the number of prompting tokens without degrading the quality of the response generated compared to a solution using RAG. For that, CRAG combines clustering, summarization, and aggregation methods to shrink all external knowledge provided to some LLM to fit into the context window size of the model. We show that CRAG can reduce the number of tokens by at least 46\%, achieving more than 90\% in some cases, compared to RAG. By reducing the number of tokens, CRAG allows a significant cost and latency reduction. For future work, alternative clustering algorithms other than K-means might be explored. Also, we opted for Mistral 7B to perform summarization, however, other larger open-source LLMs, such as Mixtral8x7B and Llama-2, potentially would perform better in extracting the key points, though they demand more resources. Finally, we use one-shot prompting for summarization, but a fine-tuning approach might help to improve the model's performance in this task.

\bibliographystyle{unsrtnat}
\bibliography{references}  

\begin{thebibliography}{11}
\providecommand{\natexlab}[1]{#1}
\providecommand{\url}[1]{\texttt{#1}}
\expandafter\ifx\csname urlstyle\endcsname\relax
  \providecommand{\doi}[1]{doi: #1}\else
  \providecommand{\doi}{doi: \begingroup \urlstyle{rm}\Url}\fi

\bibitem[Lewis et~al.(2020)Lewis, Perez, Piktus, Petroni, Karpukhin, Goyal, K{\"u}ttler, Lewis, Yih, Rockt{\"a}schel, et~al.]{lewis2020retrieval}
Patrick Lewis, Ethan Perez, Aleksandra Piktus, Fabio Petroni, Vladimir Karpukhin, Naman Goyal, Heinrich K{\"u}ttler, Mike Lewis, Wen-tau Yih, Tim Rockt{\"a}schel, et~al.
\newblock Retrieval-augmented generation for knowledge-intensive nlp tasks.
\newblock \emph{Advances in Neural Information Processing Systems}, 33:\penalty0 9459--9474, 2020.

\bibitem[Anand et~al.(2023)Anand, Goel, Hira, Buldeo, Kumar, Verma, Gupta, and Shah]{anand2023sciphyrag}
Avinash Anand, Arnav Goel, Medha Hira, Snehal Buldeo, Jatin Kumar, Astha Verma, Rushali Gupta, and Rajiv~Ratn Shah.
\newblock Sciphyrag-retrieval augmentation to improve llms on physics q \&a.
\newblock In \emph{International Conference on Big Data Analytics}, pages 50--63. Springer, 2023.

\bibitem[Wang et~al.(2023)Wang, Ma, and Chen]{wang2023augmenting}
Yubo Wang, Xueguang Ma, and Wenhu Chen.
\newblock Augmenting black-box llms with medical textbooks for clinical question answering.
\newblock \emph{arXiv preprint arXiv:2309.02233}, 2023.

\bibitem[Mansurova et~al.(2023)Mansurova, Nugumanova, and Makhambetova]{mansurova2023development}
Aigerim Mansurova, Aliya Nugumanova, and Zhansaya Makhambetova.
\newblock Development of a question answering chatbot for blockchain domain.
\newblock \emph{Scientific Journal of Astana IT University}, pages 27--40, 2023.

\bibitem[Reimers and Gurevych(2019{\natexlab{a}})]{reimers-2019-sentence-bert}
Nils Reimers and Iryna Gurevych.
\newblock Sentence-bert: Sentence embeddings using siamese bert-networks.
\newblock In \emph{Proceedings of the 2019 Conference on Empirical Methods in Natural Language Processing}. Association for Computational Linguistics, 11 2019{\natexlab{a}}.
\newblock URL \url{http://arxiv.org/abs/1908.10084}.

\bibitem[Song et~al.(2020)Song, Tan, Qin, Lu, and Liu]{song2020mpnet}
Kaitao Song, Xu~Tan, Tao Qin, Jianfeng Lu, and Tie-Yan Liu.
\newblock Mpnet: Masked and permuted pre-training for language understanding, 2020.

\bibitem[Reimers and Gurevych(2019{\natexlab{b}})]{reimers2019sentencebert}
Nils Reimers and Iryna Gurevych.
\newblock Sentence-bert: Sentence embeddings using siamese bert-networks, 2019{\natexlab{b}}.

\bibitem[Tan et~al.(2018)Tan, Steinbach, and Kumar]{tan2018introduction}
Pang-Ning Tan, Michael Steinbach, and Vipin Kumar.
\newblock \emph{Introduction to data mining}.
\newblock Pearson Education, 2nd edition, 2018.

\bibitem[MacQueen(1967)]{macqueen1967classification}
J~MacQueen.
\newblock Classification and analysis of multivariate observations.
\newblock In \emph{Proceedings of the 5th Berkeley Symposium on Mathematical Statistics and Probability}, pages 281--297, 1967.

\bibitem[Jiang et~al.(2023)Jiang, Sablayrolles, Mensch, Bamford, Chaplot, de~las Casas, Bressand, Lengyel, Lample, Saulnier, Lavaud, Lachaux, Stock, Scao, Lavril, Wang, Lacroix, and Sayed]{jiang2023mistral}
Albert~Q. Jiang, Alexandre Sablayrolles, Arthur Mensch, Chris Bamford, Devendra~Singh Chaplot, Diego de~las Casas, Florian Bressand, Gianna Lengyel, Guillaume Lample, Lucile Saulnier, Lélio~Renard Lavaud, Marie-Anne Lachaux, Pierre Stock, Teven~Le Scao, Thibaut Lavril, Thomas Wang, Timothée Lacroix, and William~El Sayed.
\newblock Mistral 7b, 2023.

\bibitem[OpenAI and et. al(2023)]{openai2023gpt4}
OpenAI and Josh~Achiam et. al.
\newblock Gpt-4 technical report, 2023.

\end{thebibliography}






\end{document}